\documentclass[conference]{IEEEtran}
\IEEEoverridecommandlockouts
\usepackage{cite}
\usepackage{amsmath,amssymb,amsfonts}
\usepackage{algorithmic}
\usepackage{graphicx}
\usepackage{textcomp}
\usepackage{xcolor}
\usepackage{algorithmic}
\usepackage{algorithm}
\usepackage{caption}
\usepackage{caption}
\def\BibTeX{{\rm B\kern-.05em{\sc i\kern-.025em b}\kern-.08em
    T\kern-.1667em\lower.7ex\hbox{E}\kern-.125emX}}

\begin{document}

\title{Towards Chip-in-the-loop Spiking Neural Network Training via Metropolis-Hastings Sampling \\

}

\author{
\IEEEauthorblockN{Ali Safa$^{1,2}$, Vikrant Jaltare$^3$, Samira Sebt$^3$, Kameron Gano$^3$, Johannes Leugering$^3$, 
\\
Georges Gielen$^{1,2}$, Gert Cauwenberghs$^3$}
\IEEEauthorblockA{\textit{$^{1}$imec, Leuven, Belgium, $^{2}$ESAT, KU Leuven, Belgium, $^3$University of California at San Diego, La Jolla, USA}\\
Ali.Safa@imec.be}

}

\maketitle

\begin{abstract}
This paper studies the use of Metropolis-Hastings sampling for training Spiking Neural Network (SNN) hardware subject to strong unknown non-idealities, and compares the proposed approach to the common use of the backpropagation of error (backprop) algorithm and surrogate gradients, widely used to train SNNs in literature. Simulations are conducted within a chip-in-the-loop training context, where an SNN subject to unknown distortion must be trained to detect cancer from measurements, within a biomedical application context. Our results show that the proposed approach strongly outperforms the use of backprop by up to $27\%$ higher accuracy when subject to strong hardware non-idealities. Furthermore, our results also show that the proposed approach outperforms backprop in terms of SNN generalization, needing $>10 \times$ less training data for achieving effective accuracy. These findings make the proposed training approach well-suited for SNN implementations in analog subthreshold circuits and other emerging technologies where unknown hardware non-idealities can jeopardize backprop.
\end{abstract}


\vspace{-1mm}

\section{Introduction}

In recent years, Spiking Neural Networks (SNNs) \cite{roadmap} have attracted much attention for extreme-edge AI applications such as healthcare monitoring, wearables and other battery-power IoT applications where machine learning tasks must be run with the tightest of energy, latency and area budgets \cite{biomedsnn, digitalexamle, example2, biomedsnn2, texture}. Compared to conventional Deep Neural Networks (DNNs) \cite{demo}, SNNs make use of event-driven binary activations (or spikes), closely replicating how biological neurons generate spikes in the brain \cite{stdptheo}. This spiking, event-driven nature, makes SNNs highly appropriated for ultra-low-power applications since SNNs only consume dynamic energy when a spike is emitted \cite{timedomneuron}. This significantly reduces power consumption when implemented in neuromorphic hardware, compared to frame-based DNNs processing continuous values \cite{eprop_chip, cim_learn, cim_chip}. In addition, their binary activation nature drops the need for costly weight multiplication hardware since inner products $\Bar{W}^T \Bar{s}$ between weight vectors $\Bar{W}$ and binary spike vectors $\Bar{s}$ can be reduced to a simple addition of the weight elements corresponding to a spiking entry in $\Bar{s}$, further reducing energy and area overheads in neuromorphic hardware \cite{microiee, reram}. 

In addition to the hardware-efficient working properties of SNNs, many teams are currently exploring the implementation of SNNs in analog sub-threshold hardware instead of less power-efficient digital solutions \cite{analogexamle, analogexamle2, analogneuron, memrist}. But an additional challenge that comes with analog sub-threshold SNN designs is the sensitivity of the underlying hardware to Process, Voltage, and Temperature (PVT) variations, as well as distortions, hysteresis, and non-linearities arising from the biasing of transistors in the subthreshold regime \cite{pvt1, pvt2}. Due to the dependency of these non-idealities to PVT variations, it is often hard to build \textit{accurate software models} of the obtained SNN chip since its properties can vary  both with time and across different produced chips \cite{nips}. 

Still, current SNN training approaches such as the widely-used backpropagation of error (backprop) algorithm coupled with the use of surrogate gradients \cite{snnradarsurrogate} (for alleviating the ill-defined nature of the Dirac spike derivative) \textit{do require} the availability of a software model describing the behavior of each spiking neuron in the hardware \cite{lessonssnn, friedman}. This jeopardizes the direct deployment of an SNN model trained offline in analog SNN implementations, where an accurate model of each chip is not always available. Indeed, since the underlying hardware does not precisely match the SNN software model, a direct deployment of the trained model would result in significant losses in accuracy \cite{memristor_mcmc}. 
\begin{figure}[t]
\centering
      \includegraphics[scale = 0.42]{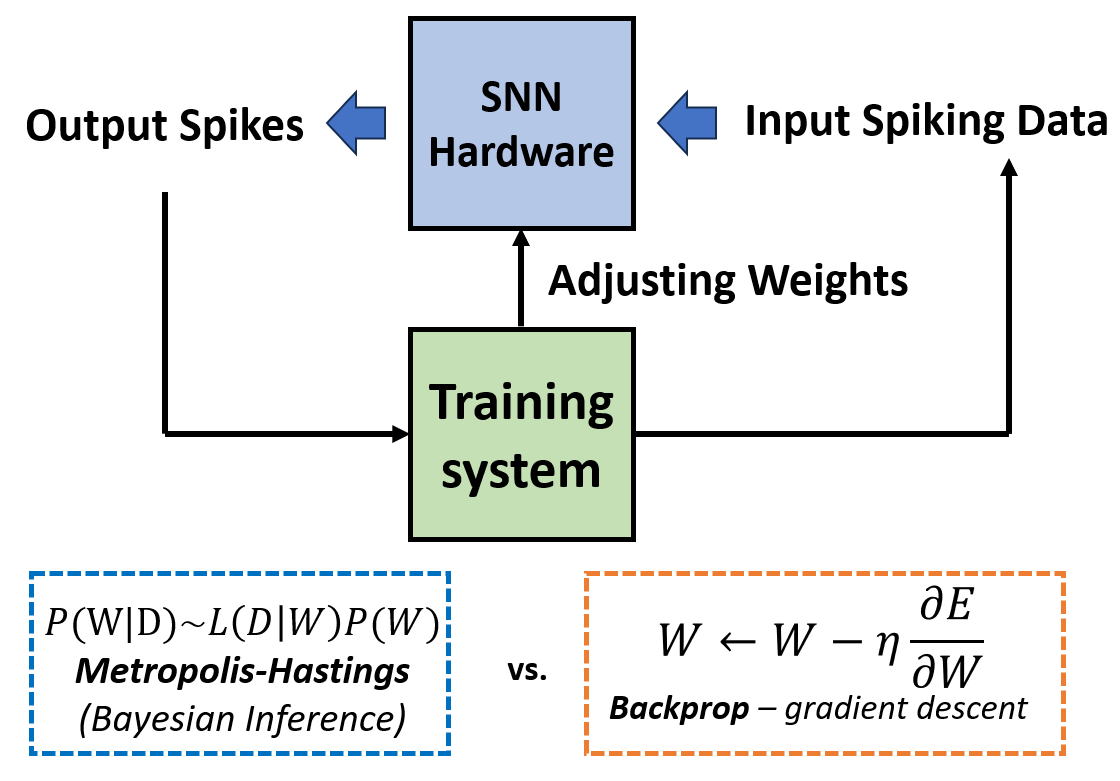}
      \caption{\textit{\textbf{Chip-in-the-loop training of SNN hardware.} This work studies Metropolis-Hastings sampling, where the SNN weights $W$ are learned (inferred) using Bayes' rule with $P(W|D), P(W)$ the posterior and prior densities, and $L(D|W)$ the likelihood density given the data $D$ (see Section \ref{mhdescription}). The proposed method is compared to conventional back-propagation of error, where the weights $W$ are learned via gradient descent for minimizing the error $E$. 
      }}
      \label{chipintheloop}
 \end{figure}
 
To alleviate this issue, chip-in-the-loop training approaches \cite{hwinloop, brainscales} (see Fig. \ref{chipintheloop}) have been proposed where the actual SNN hardware is used for the \textit{inference} (or forward propagation) pass of the training procedure, and the inference result is then used to evaluate the error function ($E$ in Fig. \ref{chipintheloop}) and perform backprop accordingly. But still, these approaches rely on the \textit{assumption} that the surrogate gradients used during chip-in-the-loop training are still valid gradient approximations even in the presence of unknown SNN non-idealities, which might not always be the case in current and future analog implementations, as the limits of energy and area efficiency are pushed further \cite{roadmap, cim_chip, memory_crossbar}.

Hence, it is crucial to investigate \textit{model-free} approaches (such as \textit{Bayesian Inference} techniques) for SNN training, which do not require the use of the surrogate gradient method and the assumptions that it entails. To this end, this paper proposes a novel approach for training SNNs via \textit{Metropolis-Hastings} sampling \cite{metropolishastings} (which performs the inference of the SNN weights $W$ given the input data $D$ following Bayes' rule in Fig. \ref{chipintheloop}), and motivate its use for SNN hardware subject to strong, unknown non-idealities.  

The main contributions of this paper are the followings:
\begin{enumerate}
    \item We propose a novel approach for training SNNs via Metropolis-Hastings sampling \cite{metropolishastings} for chip-in-the-loop training of SNN hardware subject to potential distortion and non-idealities, such as in analog sub-threshold SNN implementations.
    \item We show that our proposed Metropolis-Hastings SNN training approach significantly outperforms the use of conventional surrogate gradient backprop in terms of accuracy when the SNN hardware is subject to unknown distortion non-idealities.
    \item We also show that our proposed Metropolis-Hastings SNN training approach enables the SNN to generalize significantly better to unknown data during test inference, remarkably needing more than one order of magnitude \textit{less training data} in order to achieve similar or higher accuracy compared to backprop.
\end{enumerate}

This paper is organized as follows. Section \ref{methods_sec} introduces our proposed Metropolis-Hastings SNN training method for chip-in-the-loop learning in SNN hardware subject to strong non-linear distortion and defects. Section \ref{result_sec} shows our results and comparison against conventional surrogate gradient backprop training. Finally, conclusions are provided in Section \ref{conc_sec}.
\section{Methods}
\label{methods_sec}

\subsection{Chip-in-the-loop SNN training via Metropolis-Hastings}
\label{mhdescription}

In this paper, we propose to use Metropolis-Hastings sampling \cite{metropolishastings} as a Bayesian method for training the SNN weights, without the need for a precise model detailing the underlying SNN hardware and its non-idealities. Metropolis-Hastings provides a computationally tractable way for estimating the posterior distribution of the SNN weights $P(W|D)$ given the data $D$ following Bayes' rule \cite{metropolishastings}:
\begin{equation}
    P(W_{n+1}|D) = \frac{L(D|W_n) P(W_n)}{P(D)}
    \label{bayes}
\end{equation}
where $n$ is the current iteration of the Bayesian inference process, $W_n$ denotes the SNN weights at step $n$, $L(D|W_n)$ is the likelihood probability distribution, $P(W_n)$ is the prior distribution, $P(D)$ is the data distribution and $P(W_{n+1}|D)$ the posterior distribution of $W$ at step $n+1$. 

After the convergence of the iterative process (\ref{bayes}), the SNN weights $W^*$ are obtained as the \textit{a posteriori} mean value \cite{map_bayes}: 
\begin{equation}
    W^* = \int W^* P(W^*|D) dW^*
    \label{map}
\end{equation}


The rest of this section describes the definition of the probability density functions in (\ref{bayes}), followed by a description of how each probability density is used by the Metropolis-Hastings algorithm for training the SNN in Algorithm \ref{mhsamp}. 
\vspace{1pt}
\subsubsection{Likelihood density model}
\label{likelihood_comp}
We consider an arbitrary SNN architecture with $n$ layers and with the last layer working as a \textit{classification} output layer, with as much output leaky integrate and fire (LIF) neurons as the number of classes in the task to be solved. We denote $W$ as the tensor containing all the SNN weights.

To compute the Likelihood density, we first feed the training input data $D_i, \forall i = 1,...,N_{train}$ to the SNN which produces output spikes $\Bar{s}_{out}[k]$ (where $i$ denotes the index of the data point in the train set and $k$ denotes the time step).

Since the goal is to steer the spiking rate of the output classification layer to correspond to the one-hot encoded class labels in the train set, we compute the output spiking rate as:
\begin{equation}
    \Bar{r}_{out,i} = \frac{1}{T} \sum_{k=1}^{T} \Bar{s}_{out}[k]
\end{equation}
which contains values between $0$ and $1$ (with $T$ the total number of time steps).

Then, we compute the mean square error (MSE) between the output rate vector $\Bar{r}_{out,i}$ of the SNN and the one-hot encoded label $\Bar{y}_i$ corresponding to the $i^{th}$ data sample in the train set:
\begin{equation}
    MSE = \frac{1}{N_{train}} \sum_{i=1}^{N_{train}} ||\Bar{y}_i -  \Bar{r}_{out,i}||_2^2
\end{equation}
Then, the MSE is used to form a Gaussian likelihood density (since maximizing a Gaussian likelihood is similar to minimizing the MSE error, see \cite{mseaslikelihood}):
\begin{equation}
    L(D|W) = \frac{1}{ \sqrt{2 \pi} v} e^{\frac{-MSE}{2 v^2}} 
    \label{likelihood}
\end{equation}
where $v^2$ is the variance of the chosen Likelihood function.
\vspace{1pt}
\subsubsection{Prior density model}
As prior distribution for each SNN weight $W_j$, we use a Gaussian distribution with variance $p^2$:
\begin{equation}
    P(W_j) = \frac{1}{ \sqrt{2 \pi} p} e^{\frac{-W^2_j}{2 p^2}} 
    \label{prior}
\end{equation}

\subsubsection{Posterior density inference} We use Metropolis-Hastings sampling \cite{metropolishastings} (Algorithm \ref{mhsamp}) to estimate the posterior density over the weights $P(W|D)$ by using (\ref{likelihood}) as the likelihood model and (\ref{prior}) as the prior model. Metropolis-Hastings sampling is an implicit approach for maximizing the posterior without explicitly using Bayes' rule, which becomes significantly more complex to compute as the dimension of the weight tensor $W$ (i.e., the number of SNN weights) becomes important \cite{bayescoplex}. Algorithm \ref{mhsamp} describes the Metropolis-Hastings SNN training procedure.



\begin{algorithm}
 \caption{Metropolis-Hastings SNN Training}
  \label{mhsamp}
 \begin{algorithmic}[1]
 \renewcommand{\algorithmicrequire}{\textbf{Input:}}
 \renewcommand{\algorithmicensure}{\textbf{Output:}}
 \REQUIRE $D$ : training data and labels, $c^2$ : weight sampling variance, $v^2$ : Likelihood density variance in (\ref{likelihood}), $p^2$ : Prior density variance in (\ref{prior}).
 \STATE $\Lambda \xleftarrow{} c^2 I$ //diagonal covariance matrix with variance $c^2$ 
 \STATE $W_1 \sim N(0,\Lambda)$  
  \FOR {$n = 1$ to end}
  \STATE $W_{p} \sim N(W_{n}, \Lambda)$ // Sample random proposal $W_p$
  \STATE \textbf{Run SNN} with weights $W_p$ and $W_n$ and data $D$ to compute $L(D | W_{p})$ and $L(D | W_{n})$ (see Section \ref{likelihood_comp})
  \STATE $\alpha(W_{p}|W_{n}) = \min \{1, \frac{L(D | W_{p}) P(W_{p}) }{L(D | W_{n}) P(W_{n})} \}$
  \STATE With probability $\alpha(W_{p}|W_{n})$ set $W_{n+1} = W_p$, \textbf{else} keep $W_{n+1} = W_{n}$
  \ENDFOR
 \end{algorithmic} 
 \end{algorithm}

\subsection{LIF neuron with distortion defects}
In order to study the effect of potentially strong sub-threshold non-linearities and possible circuit defects causing additional non-linear behavior, we use the following modified LIF neuron model with an arbitrary non-linearity $f_\sigma$ applied to the membrane potential accumulation $V_k$:
\begin{equation}
 \begin{cases}
    V_{k+1} = \alpha V_{k} + (1 - \alpha) I_{syn} 
    \\
    V_{k+1} = f_\sigma \{ V_{k+1}\}
    \\
    S_{out} = 1 \hspace{3pt} \text{\textbf{if}} \hspace{3pt} V_{k+1} \geq \mu \hspace{3pt} \text{\textbf{else}} \hspace{3pt} 0
    \\
    V_{k+1} = 0 \hspace{3pt} \text{\textbf{if}} \hspace{3pt} V_{k+1} \geq \mu \hspace{3pt} \text{\textbf{or }} \hspace{3pt} V_{k+1} < 0
  \end{cases}
  \label{liff}
\end{equation}
where $I_{syn}$ is the input synaptic current, $\alpha$ is the membrane decay parameter, $\mu$ is the spiking threshold, $S_{out}$ is the spiking output and the non-linearity $f_\sigma$ is defined as:
\begin{equation}
    f_\sigma = x + \frac{\sigma}{2} x^2 + \frac{\sigma}{6} x^3 \approx \exp{(\sigma x)}
    \label{nonlin1}
\end{equation}
where $\sigma$ defines the strength of the non-linearity (the higher $\sigma$, the more non linear).

Crucially, the arbitrary non-linearity in (\ref{nonlin1}) is purposefully taken to benchmark the proposed MH Sampling approach for SNN training under challenging unknown distortions and defects. Therefore, using this challenging distortion model will enable our experimental results and discussions to also cover the case of \textit{less challenging} distortion models that might be found in different circuits.

This type of non-linearity can for example be found in the context of analog sub-threshold LIF circuits \cite{analogexamle, analogexamle2}, where MOSFETs biased in the sub-threshold region $V_{GS} < V_T$ 
exhibit a non-linear behavior (\ref{nonlin}) in terms of their drain current $I_D$ in function of their gate-source voltage $V_{GS}$, which is dependent on unknown Process Voltage and Temperature (PVT) variations. 
\begin{equation}
    I_D \propto \exp{(\frac{q V_{GS}}{nkT})}
    \label{nonlin}
\end{equation}
where $\frac{nkT}{q}$ is the thermal voltage. 

Hence, the sensibility of analog sub-threshold LIF neuron designs, to unknown PVT variations, serves as strong further motivation for the study of novel model-free approaches to SNN training, such as the Metropolis-Hastings sampling technique detailed in Section \ref{mhdescription}.


\section{Simulation Results}
\label{result_sec}
\subsection{SNN setup}

We conduct our experiments using the popular \textit{Wisconsin Breast Cancer} (WBC) dataset \cite{wbcdataset} for the binary detection of cancer from bio-signal measurements. The dataset features $569$ data points as vectors composed of $30$ measurement features for each data point. 

As SNN architecture, we use a \textit{1-hidden-layer} fully-connected SNN using LIF neurons subject to the non-ideality in (\ref{liff}), which becomes equivalent to a regular LIF neuron when the distortion strength $\sigma = 0$. We use \textit{$3$ neurons} in the hidden layer and \textit{$2$ output neurons} corresponding to each class in our cancer detection problem (class $1$: \textit{has cancer}, class $2$: \textit{does not have cancer}).

Before the SNN learning phase, we normalize the entries in each data vector $\Bar{x}$ to the $[0,1]$ range via $\Bar{x} \xleftarrow{} \frac{\Bar{x}}{\max \Bar{x} }$ and we perform a random $80\%/20\%$ train-test split of the complete dataset before a Poisson encoding \cite{poisson} of each data point into a vector of spike trains with $T$ time bins, where the probability of having a spike in each time bin for the $i^{th}$ entry in $\Bar{x}$ is a Bernoulli process with probability $x_i$. We arbitrary set $T=10$ to limit the time complexity of the system (the larger $T$, the longer the SNN latency and hence, the higher the power consumption when implemented in hardware).

For training the SNN, we use and compare two different learning approaches: \textit{1)} the proposed Metropolis-Hastings method described in Section \ref{mhdescription} and \textit{2)} conventional backprop using the surrogate gradient technique as the dominant approach for training SNNs in literature \cite{snnradarsurrogate, lessonssnn, friedman}. 

\subsubsection{Metropolis-Hastings parameters}
We use the Metropolis-Hastings SNN training approach described in Algorithm \ref{mhsamp}, Section \ref{mhdescription} with parameters $c = 0.5$, $v = 1$ and $p=10$ in Algorithm \ref{mhsamp}.

\subsubsection{Backprop parameters}
We train the SNN via backprop in a similar manner as in \cite{snnradarsurrogate, lessonssnn, friedman}, by using a Gaussian surrogate (\ref{surrogate}) as the replacement gradient of the LIF neuron non-linearity.
\begin{equation}
    S_{out}'(V) \approx \frac{1}{\sqrt{2 \pi}} e^{-2V^2}
    \label{surrogate}
\end{equation}
where $S_{out}$ is the spiking output and $V$ is the membrane potential (see Eq. \ref{liff}). For training, we use the Adam optimizer with learning rate $\eta=0.001$  for a total of $100$ epochs with batch size $32$.

SNN evaluation is carried via a 5-fold train-test procedure where each fold uses a different random train-test partitioning. The final test accuracy and its standard deviation are computed over the $5$ folds and reported during the experiments in Fig. \ref{fig1} and \ref{allgenres} (as the point marker and the shaded area respectively). 

\subsection{Accuracy in function of distortion strength $\sigma$}

\begin{figure}[htbp]
\centering
    \includegraphics[scale = 0.5]{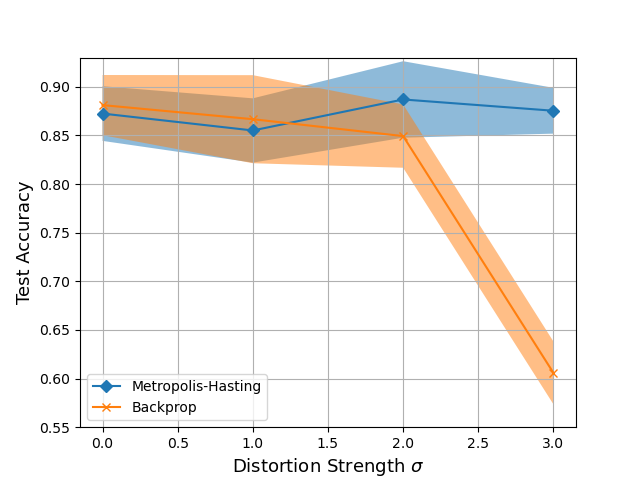}
    \caption{\textit{\textbf{Test accuracy in function of distortion strength $\sigma$.} While backprop accuracy strongly degrades when $\sigma \geq 3$, Metropolis-Hastings accuracy is not affected by the increase in LIF distortion. }}
    \label{fig1}
\end{figure}

\begin{figure*}
\centering
\begin{tabular}{cc}
  \includegraphics[width=80mm]{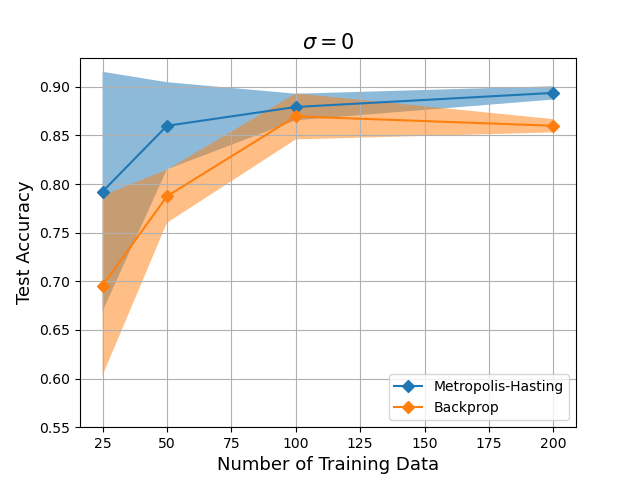} &   \includegraphics[width=80mm]{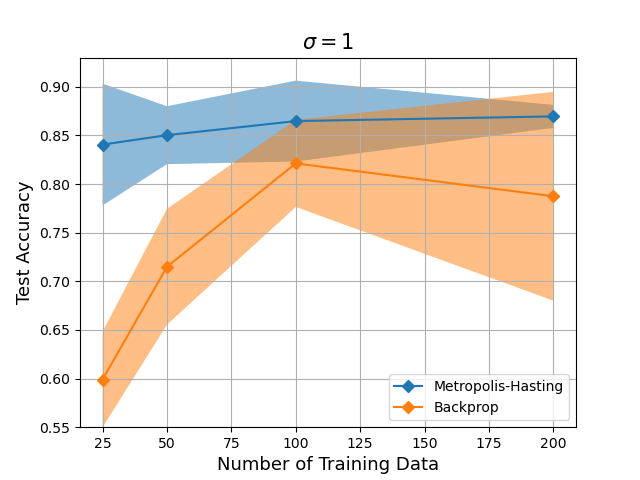} \\
(a) No distortion ($\sigma = 0$) & (b) Low distortion ($\sigma = 1$) \\[6pt]
 \includegraphics[width=80mm]{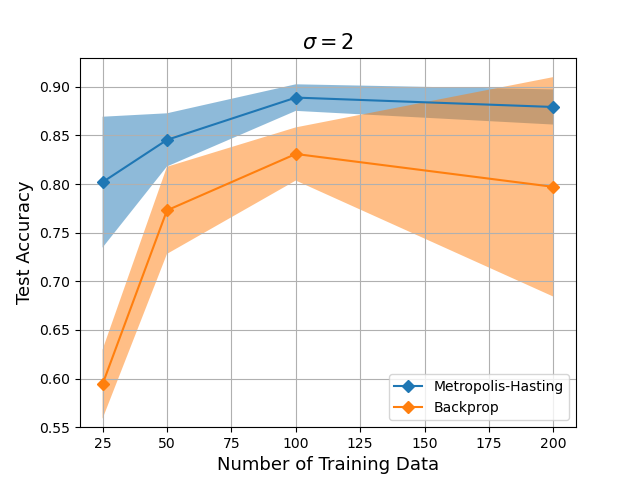} &   \includegraphics[width=80mm]{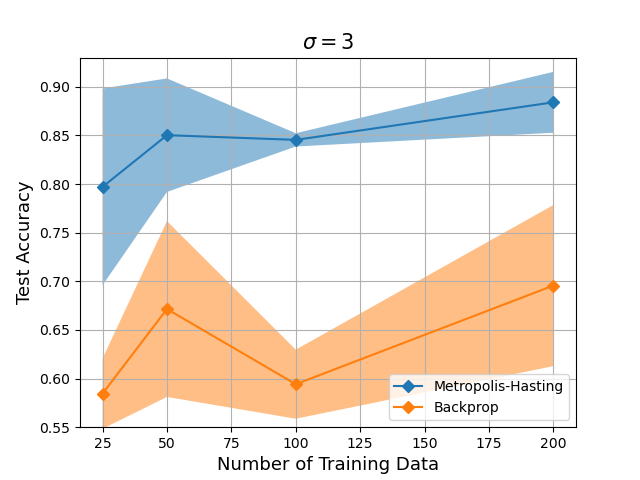} \\
(c) Mild distortion ($\sigma = 2$) & (d) Strong distortion ($\sigma = 2$) \\[6pt]
\end{tabular}
\caption{\textit{\textbf{Test accuracy in function of the train set size.}}}
\label{allgenres}
\end{figure*}

In this Section, we compare the test accuracy obtained via the proposed Metropolis-Hastings approach against the use of conventional surrogate gradient backprop, as a function of the distortion strength $\sigma$. It can be clearly seen in Fig. \ref{fig1} that the test accuracy achieved by the proposed Metropolis-Hastings sampling is significantly higher than the accuracy obtained via backprop when $\sigma \geq 2$ (a gain of $+27\%$ for $\sigma = 4$ and $+4\%$ for $\sigma = 3$). On the other hand, test accuracy is similar for both methods when $\sigma < 2$, since the difference between the distorted LIF model and the ideal LIF model used by the surrogate gradient backprop training is reduced. 
\subsection{SNN generalization performance}
In this Section, we explore the generalization ability of the SNN obtained via each training approach, by studying the test accuracy as a function of the number of training data points in the train set. Indeed, achieving faster generalization by using less training data is highly attractive in chip-in-the-loop training scenarios since it will significantly reduce memory consumption and training time, thereby reducing power consumption if chip-in-the-loop training is used on-line to e.g. reconfigure SNN hardware at the extreme edge. 

Fig. \ref{allgenres} respectively show the generalization performance of each training approach for $\sigma = 0, 1, 2, 3$. It can be clearly seen on all plots that the Metropolis-Hastings training approach leads to significantly better SNN generalization compared to backprop. When no distortion is used ($\sigma = 0$ in Fig. \ref{allgenres}), the Metropolis-Hastings approach already achieves $+10\%$ vs. backprop in test accuracy when using only $25$ data points in the train set, and still achieves higher test accuracy when more training data is used. This is thought to be due to the unique characteristics that come with Baye's rule (\ref{bayes}), placing emphasis on the prior distribution as a means to alleviate the effects of uncertainty and noise. In the case of Fig. \ref{allgenres}, this uncertainty is due to a significant lack of training data. Therefore, as expected, the MH model outperforms the traditional backprop model in terms of generalization ability.    

When distortion is added for $\sigma = 1, 2, 3$, the gain in accuracy obtained via Metropolis-Hastings compared to backprop becomes even more significant (see Fig. \ref{allgenres}). These experiments clearly demonstrate the superior generalization ability of the proposed Metropolis-Hastings SNN training approach, significantly reducing the amount of training data that needs to be collected and used, reducing the training time, power consumption, and memory requirements when e.g., using chip-in-the-loop training at the extreme edge to reconfigure the SNN hardware.

\section{Conclusion}
\label{conc_sec}

This paper has provided a study of Metropolis-Hastings sampling as a novel approach for chip-in-the-loop SNN training, where the underlying SNN hardware is subject to strong distortion non-idealities. Simulation results using the Wisconsin Breast Cancer detection dataset have shown that the proposed approach strongly outperforms the use of backprop by up to $27\%$ higher accuracy under LIF neuron distortion. In addition, it was shown that the proposed approach outperforms backprop in terms of SNN generalization, needing $>10 \times$ less training data for achieving effective classification accuracy. Hence, our findings demonstrate that the proposed training approach is well-suited for SNN implementations in analog subthreshold circuits and other emerging technologies where unknown hardware non-idealities such as distortion and hysteresis can jeopardize the use of backprop.

\section*{Acknowledgment}
This research was partially funded by a Long Stay Abroad grant from the
Flemish Fund of Research - Fonds Wetenschappelijk Onderzoek (FWO) - grant
V413023N. This research received funding from the Flemish Government under
the “Onderzoeksprogramma Artificiele Intelligentie (AI) Vlaanderen” programme.


\begin{thebibliography}{00}
\bibliographystyle{unsrt}

\bibitem{roadmap} D. Christensen, R. Dittmann, B. Linares-Barranco, A. Sebastian, M. Le Gallo, A. Redaelli, S. Slesazeck, T. Mikolajick, S. Spiga, S. Menzel, I. Valov, G. Milano, C. Ricciardi, S. Liang, F. Miao, M. Lanza, T. Quill, S. Keene, A. Salleo, J. Grollier, D. Marković, A. Mizrahi, P. Yao, J. Yang, G. Indiveri, J. P. Strachan, S. Datta, E. Vianello, A. Valentian, J. Feldmann, X. Li, W. Pernice, H. Bhaskaran, S. Furber, E. Neftci, F. Scherr, W. Maass, S. Ramaswamy, J. Tapson, P. Panda, Y. Kim, G. Tanaka, S. Thorpe, C. Bartolozzi, T. Cleland, C. Posch, S. Liu, G. Panuccio, M. Mahmud, A. Mazumder, M. Hosseini, T. Mohsenin, E. Donati, S. Tolu, R. Galeazzi, M. Christensen, S. Holm, D. Ielmini and N. Pryds (2022). "2022 roadmap on neuromorphic computing and engineering." Neuromorphic Computing and Engineering. 2. 10.1088/2634-4386/ac4a83. 

\bibitem{biomedsnn} H. Chu, H. Jia, Y. Yan, Y. Jin, L. Qian, L? Gan, Y. Huan, L. Zheng, Z. Zou, "A Neuromorphic Processing System for Low-Power Wearable ECG Classification," 2021 IEEE Biomedical Circuits and Systems Conference (BioCAS), Berlin, Germany, 2021, pp. 1-5, doi: 10.1109/BioCAS49922.2021.9644939.

\bibitem{digitalexamle} B. U. Pedroni, S. Sheik, H. Mostafa, S. Paul, C. Augustine and G. Cauwenberghs, "Small-footprint Spiking Neural Networks for Power-efficient Keyword Spotting," 2018 IEEE Biomedical Circuits and Systems Conference (BioCAS), Cleveland, OH, USA, 2018, pp. 1-4, doi: 10.1109/BIOCAS.2018.8584832.

\bibitem{example2} C. Li, Y. Wang, J. Zhang, X. Cui and R. Huang, "A Compact and Accelerated Spike-based Neuromorphic VLSI Chip for Pattern Recognition," 2018 IEEE Biomedical Circuits and Systems Conference (BioCAS), Cleveland, OH, USA, 2018, pp. 1-4, doi: 10.1109/BIOCAS.2018.8584765.

\bibitem{biomedsnn2} T. Wang, H. Wang, J. He, Z. Zhong, F. Tang, X. Zhou, S. Yu, L. Liu, N. Wu, M. Tian, C. Shi, "MorphBungee: An Edge Neuromorphic Chip for High-Accuracy On-Chip Learning of Multiple-Layer Spiking Neural Networks," 2022 IEEE Biomedical Circuits and Systems Conference (BioCAS), Taipei, Taiwan, 2022, pp. 255-259, doi: 10.1109/BioCAS54905.2022.9948539.

\bibitem{texture} H. Al Haj Ali, A. Dabbous, A. Ibrahim and M. Valle, "Spiking Neural Network Based on Threshold Encoding For Texture Recognition," 2022 29th IEEE International Conference on Electronics, Circuits and Systems (ICECS), Glasgow, United Kingdom, 2022, pp. 1-4, doi: 10.1109/ICECS202256217.2022.9971099.

\bibitem{demo} Z. Zhong, T. Wang, H. Wang, Z. Zhou, J. He, F. Tang, X. Zhou, S. Yu, L. Liu, N. Wu, M. Tian, C. Shi, "Live Demonstration: Face Recognition at The Edge Using Fast On-Chip Deep Learning Neuromorphic Chip," 2023 IEEE 5th International Conference on Artificial Intelligence Circuits and Systems (AICAS), Hangzhou, China, 2023, pp. 1-2, doi: 10.1109/AICAS57966.2023.10168667.

\bibitem{stdptheo} A. Safa et al., "Fusing Event-based Camera and Radar for SLAM Using Spiking Neural Networks with Continual STDP Learning," 2023 IEEE International Conference on Robotics and Automation (ICRA), London, United Kingdom, 2023, pp. 2782-2788, doi: 10.1109/ICRA48891.2023.10160681.

\bibitem{timedomneuron} J. Song, J. Shirn, H. Kim and W. -S. Choi, "Energy-Efficient High-Accuracy Spiking Neural Network Inference Using Time-Domain Neurons," 2022 IEEE 4th International Conference on Artificial Intelligence Circuits and Systems (AICAS), Incheon, Korea, Republic of, 2022, pp. 5-8, doi: 10.1109/AICAS54282.2022.9870009.

\bibitem{eprop_chip} Bohnstingl T., Surina A., Fabre M., Demirag, Y., Frenkel C., Payvand M., Indiveri G., Pantazi A., "Biologically-inspired training of spiking recurrent neural networks with neuromorphic hardware," 2022 IEEE 4th International Conference on Artificial Intelligence Circuits and Systems (AICAS), Incheon, Korea, Republic of, 2022, pp. 218-221, doi: 10.1109/AICAS54282.2022.9869963.

\bibitem{cim_learn} Z. Zhao, Y. Wang, X. Zhang, X. Cui and R. Huang, "An Energy-Efficient Computing-in-Memory Neuromorphic System with On-Chip Training," 2019 IEEE Biomedical Circuits and Systems Conference (BioCAS), Nara, Japan, 2019, pp. 1-4, doi: 10.1109/BIOCAS.2019.8918995.

\bibitem{cim_chip} Weier Wan, Rajkumar Kubendran, Clemens Schaefer, Sukru Burc Eryilmaz, Wenqiang Zhang, Dabin Wu, Stephen Deiss, Priyanka Raina, He Qian, Bin Gao, Siddharth Joshi, Huaqiang Wu, H-S Philip Wong, Gert Cauwenberghs "A compute-in-memory chip based on resistive random-access memory." Nature 608, 504–512 (2022). https://doi.org/10.1038/s41586-022-04992-8

\bibitem{memory_crossbar} T. Bohnstingl, A. Pantazi and E. Eleftheriou, "Accelerating Spiking Neural Networks using Memristive Crossbar Arrays," 2020 27th IEEE International Conference on Electronics, Circuits and Systems (ICECS), Glasgow, UK, 2020, pp. 1-4, doi: 10.1109/ICECS49266.2020.9294933.

\bibitem{microiee} A. Safa, J. Van Assche, M. D. Alea, F. Catthoor and G. G. E. Gielen, "Neuromorphic Near-Sensor Computing: From Event-Based Sensing to Edge Learning," in IEEE Micro, vol. 42, no. 6, pp. 88-95, 1 Nov.-Dec. 2022, doi: 10.1109/MM.2022.3195634.

\bibitem{reram} P. Li, C. Yang, W. Chen, J. Huang, W. Wei, J. Liu, W. Lin, T. Hsu, C. Hsieh, R. Liu, M. Chang, K. Tang, "A Neuromorphic Computing System for Bitwise Neural Networks Based on ReRAM Synaptic Array," 2018 IEEE Biomedical Circuits and Systems Conference (BioCAS), Cleveland, OH, USA, 2018, pp. 1-4, doi: 10.1109/BIOCAS.2018.8584810.

\bibitem{analogexamle} Zhitao Yang, Yucong Huang, Jianghan Zhu, and Terry Tao Ye. 2020. Analog Circuit Implementation of LIF and STDP Models for Spiking Neural Networks. In Proceedings of the 2020 on Great Lakes Symposium on VLSI (GLSVLSI '20). Association for Computing Machinery, New York, NY, USA, 469–474. https://doi.org/10.1145/3386263.3406940

\bibitem{analogexamle2} A. Rubino, M. Cartiglia, M. Payvand and G. Indiveri, "Neuromorphic analog circuits for robust on-chip always-on learning in spiking neural networks," 2023 IEEE 5th International Conference on Artificial Intelligence Circuits and Systems (AICAS), Hangzhou, China, 2023, pp. 1-5, doi: 10.1109/AICAS57966.2023.10168620.

\bibitem{analogneuron} K. Yue and A. C. Parker, "Analog Neurons with Dopamine-Modulated STDP," 2019 IEEE Biomedical Circuits and Systems Conference (BioCAS), Nara, Japan, 2019, pp. 1-4, doi: 10.1109/BIOCAS.2019.8919047.

\bibitem{memrist} D. Wang, J. Xu, F. Li, L. Zhang, Y. Wang, A. Lansner, A. Hemani, L. Zheng, Z. Zou, "Memristor-Based In-Circuit Computation for Trace-Based STDP," 2022 IEEE 4th International Conference on Artificial Intelligence Circuits and Systems (AICAS), Incheon, Korea, Republic of, 2022, pp. 1-4, doi: 10.1109/AICAS54282.2022.9870015.

\bibitem{pvt1} P. Livi and G. Indiveri, "A current-mode conductance-based silicon neuron for address-event neuromorphic systems," 2009 IEEE International Symposium on Circuits and Systems, Taipei, Taiwan, 2009, pp. 2898-2901, doi: 10.1109/ISCAS.2009.5118408.

\bibitem{pvt2} Dorzhigulov A, Saxena V. "Spiking CMOS-NVM mixed-signal neuromorphic ConvNet with circuit- and training-optimized temporal subsampling." Front Neurosci. 2023 Jul 18;17:1177592. doi: 10.3389/fnins.2023.1177592. PMID: 37534034; PMCID: PMC10390782.

\bibitem{nips} Qu Yang, Jibin Wu, Malu Zhang, Yansong Chua, Xinchao Wang, Haizhou Li (2022). "Training Spiking Neural Networks with Local Tandem Learning." In Advances in Neural Information Processing Systems.

\bibitem{snnradarsurrogate} A. Safa et al., "Improving the Accuracy of Spiking Neural Networks for Radar Gesture Recognition Through Preprocessing," in IEEE Transactions on Neural Networks and Learning Systems, vol. 34, no. 6, pp. 2869-2881, June 2023, doi: 10.1109/TNNLS.2021.3109958.

\bibitem{lessonssnn} J. K. Eshraghian, M. Ward, E. Neftci, X. Wang, G. Lenz, G. Dwivedi, M. Bennamoun, D. S. Jeong, W. D. Lu, "Training Spiking Neural Networks Using Lessons From Deep Learning," in Proceedings of the IEEE, vol. 111, no. 9, pp. 1016-1054, Sept. 2023, doi: 10.1109/JPROC.2023.3308088.

\bibitem{friedman} E. O. Neftci, H. Mostafa and F. Zenke, "Surrogate Gradient Learning in Spiking Neural Networks: Bringing the Power of Gradient-Based Optimization to Spiking Neural Networks," in IEEE Signal Processing Magazine, vol. 36, no. 6, pp. 51-63, Nov. 2019, doi: 10.1109/MSP.2019.2931595.

\bibitem{memristor_mcmc} Dalgaty, T., Castellani, N., Turck, C., K. Harabi, D. Querlioz, E. Vianello., "In situ learning using intrinsic memristor variability via Markov chain Monte Carlo sampling." Nat Electron 4, 151–161 (2021). https://doi.org/10.1038/s41928-020-00523-3

\bibitem{hwinloop} J. Parker Mitchell and Catherine Schuman. 2021. "Low Power Hardware-In-The-Loop Neuromorphic Training Accelerator." In International Conference on Neuromorphic Systems 2021 (ICONS 2021). Association for Computing Machinery, New York, NY, USA, Article 4, 1–4. https://doi.org/10.1145/3477145.3477150

\bibitem{brainscales} S. Schmitt, J. Klaehn, G. Bellec, A. Gruebl, M. Guettler, A. Hartel, S. Hartmann, D. Husmann, K. Husmann, V. Karasenko, M. Kleider, C. Koke, C. Mauch, E. Mueller, P. Mueller, J. Partzsch, M. A. Petrovici, S. Schiefer, S. Scholze, B. Vogginger, R. Legenstein, W. Maass, C. Mayr, J. Schemmel, K. Meier, "Neuromorphic hardware in the loop: Training a deep spiking network on the BrainScaleS wafer-scale system," 2017 International Joint Conference on Neural Networks (IJCNN), Anchorage, AK, USA, 2017, pp. 2227-2234, doi: 10.1109/IJCNN.2017.7966125.

\bibitem{metropolishastings} Robert, C.P. (2015). "The Metropolis–Hastings Algorithm." In Wiley StatsRef: Statistics Reference Online (eds N. Balakrishnan, T. Colton, B. Everitt, W. Piegorsch, F. Ruggeri and J.L. Teugels). https://doi.org/10.1002/9781118445112.stat07834

\bibitem{map_bayes} Bassett, R., Deride, J. "Maximum a posteriori estimators as a limit of Bayes estimators." Math. Program. 174, 129–144 (2019). https://doi.org/10.1007/s10107-018-1241-0

\bibitem{bayescoplex} Gregory F. Cooper, "The computational complexity of probabilistic inference using bayesian belief networks," Artificial Intelligence, Volume 42, Issues 2–3, 1990, Pages 393-405, ISSN 0004-3702, https://doi.org/10.1016/0004-3702(90)90060-D.

\bibitem{mseaslikelihood} Abt, Markus. “Estimating the Prediction Mean Squared Error in Gaussian Stochastic Processes with Exponential Correlation Structure.” Scandinavian Journal of Statistics, vol. 26, no. 4, 1999, pp. 563–78. JSTOR, http://www.jstor.org/stable/4616579. Accessed 25 Dec. 2023.

\bibitem{wbcdataset}  Wolberg, W., Mangasarian, O., Street, N., and Street, W.. (1995). "Breast Cancer Wisconsin (Diagnostic)." UCI Machine Learning Repository. https://doi.org/10.24432/C5DW2B.

\bibitem{poisson} Auge, D., Hille, J., Mueller, E., Knoll, A., A Survey of Encoding Techniques for Signal Processing in Spiking Neural Networks. Neural Process Lett 53, 4693–4710 (2021). https://doi.org/10.1007/s11063-021-10562-2






\end{thebibliography}
\end{document}